%% file: arxiv.tex
\documentclass[10pt,twocolumn,letterpaper]{article}

\usepackage{iccv}
\usepackage{times}
\usepackage{epsfig}
\usepackage{graphicx}
\usepackage{amsmath}
\usepackage{amssymb}
\usepackage{caption}

\usepackage[breaklinks=true,bookmarks=false]{hyperref}
\iccvfinalcopy 

\def\iccvPaperID{9407} 

\mathchardef\mhyphen="2D
\ificcvfinal\fi

\begin{document}

\title{DreamPose: Fashion Image-to-Video Synthesis via Stable Diffusion}

\author{Johanna Karras$^1$, Aleksander Holynski$^{2,3}$, Ting-Chun Wang$^4$, Ira Kemelmacher-Shlizerman$^1$\\ \\
$^1$University of Washington, $^2$UC Berkeley, $^3$Google Research, $^4$NVIDIA\\
}

\twocolumn[{%
\renewcommand\twocolumn[1][]{#1}%
\maketitle
\input{Figures/teaser}
    }]
\maketitle

\begin{figure*}[h!]
      \begin{center}
         \includegraphics[width=0.9\linewidth]{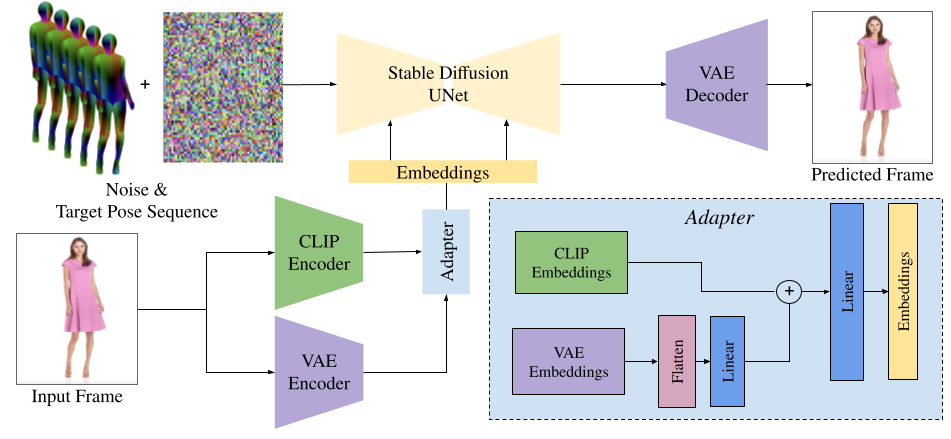}
            \caption{ Architecture Overview. We modify the original Stable Diffusion architecture in order to enable image and pose conditioning. First, we replace the CLIP text encoder with a dual CLIP-VAE image encoder and adapter module (shown in the blue box). The adapter module jointly models and reshapes the pretrained CLIP and VAE input image embeddings. Then, we concatenate the target pose representation, consisting of 5 consecutive poses surrounding the target pose, to the input noise. During training, we fine-tune the denoising UNet and our Adapter module on the full dataset and further perform subject-specific finetuning of the UNet, Adapter, and VAE decoder on a single input image.  }
        \label{fig:architecture}
        \end{center}
    \end{figure*}
    \nopagebreak
    
\vspace{-1em}
\begin{abstract}
\vspace{-1em}
We present DreamPose, a diffusion-based method for generating animated fashion videos from still images. Given an image and a sequence of human body poses, our method synthesizes a video containing both human and fabric motion. To achieve this, we transform a pretrained text-to-image model (Stable Diffusion \cite{stable_diffusion}) into a pose-and-image guided video synthesis model, using a novel finetuning strategy, a set of architectural changes to support the added conditioning signals, and techniques to encourage temporal consistency. We fine-tune on a collection of fashion videos from the UBC Fashion dataset \cite{ubc_fashion}. We evaluate our method on a variety of clothing styles and poses, and demonstrate that our method produces state-of-the-art results on fashion video animation. Video results are available on our \href{https://grail.cs.washington.edu/projects/dreampose/}{project page}: https://grail.cs.washington.edu/projects/dreampose
\end{abstract}

\vspace{-2em}
\section{Introduction}
 Fashion photography is incredibly prevalent online, from social media platforms to online retail sites. Unfortunately, these still photographs are limited in the information they convey, and fail to capture many of the crucial nuances of a garment, such as how it drapes and flows when worn. Fashion \emph{videos}, on the other hand, do showcase all these details, and for this reason are highly informative for consumer decision-making. Despite this benefit, however, these videos are a relatively rare commodity.

In this paper, we introduce DreamPose, a method that turns fashion photographs into realistic, animated videos, using a driving pose sequence. Our method is a diffusion video synthesis model based upon Stable Diffusion~\cite{stable_diffusion}. Given one or more images of a human and a pose sequence, DreamPose generates a high-quality video of the input subject following the pose sequence (Figure~\ref{fig:teaser}).
 
This is a challenging task in several ways. While image diffusion models have shown impressive, high-quality results \cite{stable_diffusion,dalle2,imagen}, video diffusion models have yet to achieve the same quality of results and are often limited to ``textural" motion or cartoon-like appearance \cite{flexible-diffusion-modeling,imagen_video,video-diffusion-models,make-a-video,ddpm_video}. Moreover, existing video diffusion models suffer from poor temporal consistency, motion jitter, lack of realism, and the inability to control the motion or detailed object appearance in the target video. This is partly because existing models are primarily conditioned on text, as opposed to other conditioning signals (e.g., motion) which may offer more fine-grained control. In contrast, our image-and-pose conditioning scheme allows for greater appearance fidelity and frame-to-frame consistency. 

Our model is fine-tuned from an existing pretrained image diffusion model, which already effectively models the distribution of natural images. When using such a model, the task of image animation can effectively be simplified to 
finding the subspace of natural images consistent with the conditioning signals. To accomplish this, we redesign the encoder and conditioning mechanisms of the Stable Diffusion~\cite{stable_diffusion} architecture, in order to enable aligned-image and unaligned-pose conditioning. Further, we propose a two-stage finetuning scheme that consists of finetuning both UNet and VAE from one or more input images.

To summarize, our contributions include: (1) \emph{DreamPose}: an image-and-pose conditioned diffusion method for still fashion image animation that achieves photorealistic results on a diverse range of garment patterns and shapes, (2) a simple, yet effective, pose conditioning approach that greatly improves temporal consistency across frames, (3) a split CLIP-VAE encoder that increases the output fidelity to the conditioning image, (4)
a finetuning strategy that effectively balances image fidelity and generalization to new poses. 
        

    \begin{figure*}[h!]
      \begin{center}
         \includegraphics[width=1.0\linewidth]{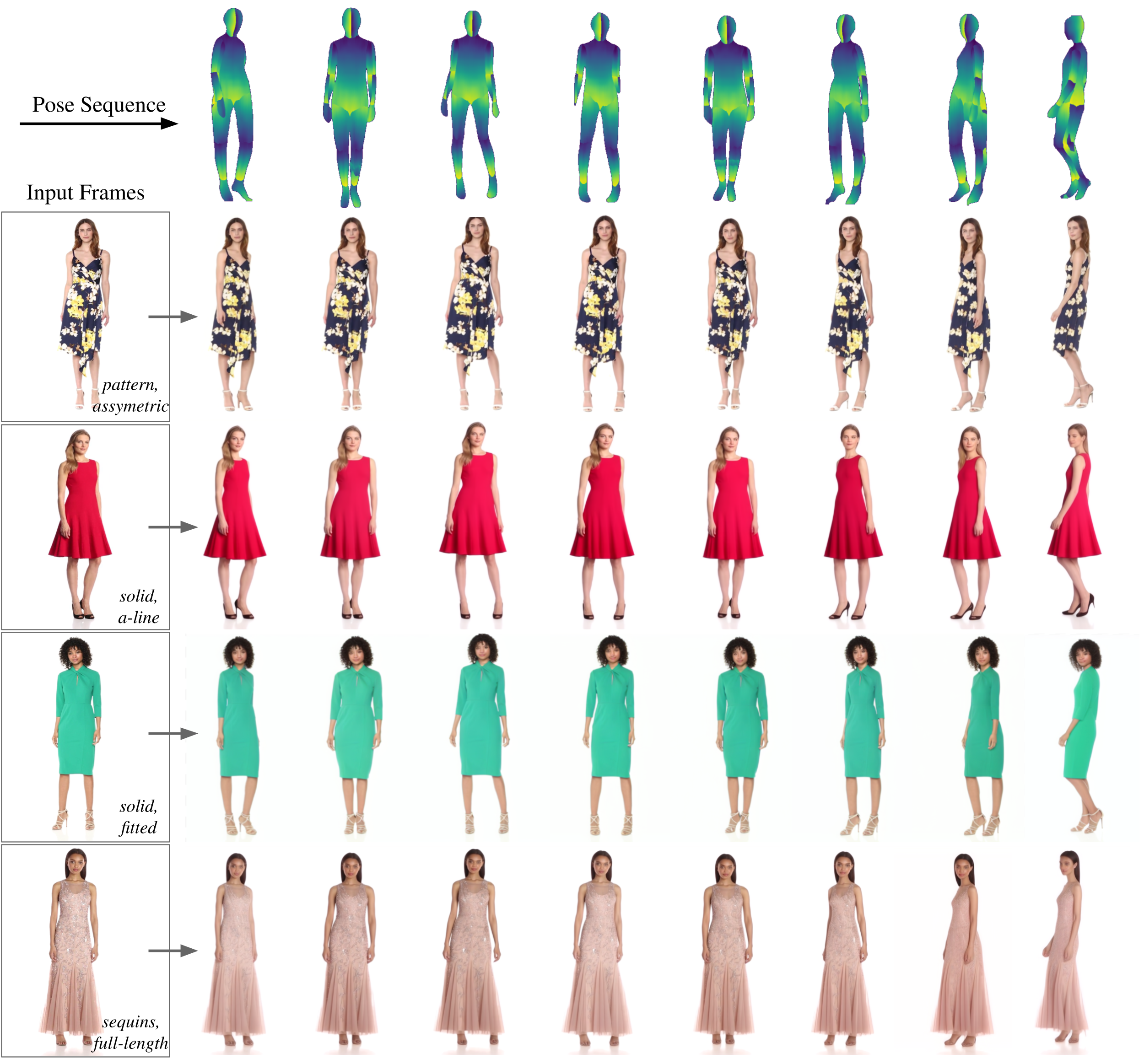}
      \end{center}
       \caption{ Qualitative Results. We showcase the results of our method on a variety of input frames and poses. DreamPose is capable of synthesizing photorealistic video frames consistent with a diverse range of patterns, fabric types, person identities, clothing shapes, and viewpoints. }
    \label{fig:qualitative-results}
    \end{figure*}
    \nopagebreak
    
\vspace{-0.5em}
\section{Related Work}
    \subsection{Diffusion models}
        Diffusion models have recently demonstrated impressive results in text-conditioned image synthesis \cite{stable_diffusion,dalle2,imagen}, video synthesis \cite{imagen_video,video-diffusion-models,dreamix}, and 3D generation tasks \cite{dream_fusion,make-a-video-3d}. However, training these models from scratch is computationally expensive and data intensive.
        Latent Diffusion Models (as in Stable Diffusion~\cite{stable_diffusion}) perform diffusion and denoising in the latent space, thereby drastically reducing the computational requirements and training time with only marginal reductions to quality.
        Since its release, Stable Diffusion and its pretrained checkpoints have been used by many for various image generation tasks \cite{universal_guidance,instruct_pix2pix,dreambooth}. Like these methods, our work leverages a pretrained Stable Diffusion model with subject-specific finetuning. 
            
    \subsection{Still Image Animation}
        Still image animation refers to the task of generating a video from one or more input images. Existing, non-diffusion approaches often consist of multiple separate networks, such as for predicting the background \cite{articulated_animation,photo_wake-up,thin_plate_spline}, motion representation \cite{eulerian_motion,monkey-net,fomm,articulated_animation,latent_image_animator,thin_plate_spline}, occlusion maps \cite{fomm,articulated_animation,photo_wake-up,thin_plate_spline}, or depth maps \cite{dain}. 
        However, multiple networks require separate training for each stage and potentially unavailable or imperfect ground-truth intermediate data, such as motion or depth. Especially with large and complex motion, these ground-truth estimates are harder to derive and more error-prone. Several more recent papers explore end-to-end single-network approaches, such as by merging optical flow and warping ~\cite{film}, replacing motion estimation networks entirely with a cross-attention modules \cite{implicit_warping}, or generating animatable 3D humans using a NeRF representation ~\cite{eva3d}. 

    \subsection{Fashion Image Synthesis}
        Many prior pose-guided fashion image synthesis methods are generative adversarial network (GAN)-based and rely on optical flow to align image features to pose \cite{pose_with_style,DiOr,try_on_gan,global_flow_local_attn,pise,progressive_pose_transfer}. 
        However, GAN-based approaches often struggle with large pose changes, synthesizing occluded regions, and preserving garment style. More recent approaches rely on attention-based mechanisms, where self- and cross-attention are used to warp image features to the target frame \cite{viton_hd,DynaST,global_flow_local_attn}. 

        Relatively few works exist for diffusion-based fashion image and video synthesis. DiffFashion~\cite{DiffFashion} aims to edit a clothing item by transferring the style of a reference image. Concurrent work PIDM~\cite{pose-transfer-dm} generates pose-conditioned human images, but is designed for single-image pose transfer, and therefore not optimized for temporal consistency. Our approach uses a unique finetuning and multi-pose input representation to enhance temporal smoothness. Moreover, by leveraging pretrained Stable Diffusion \cite{stable_diffusion}, our method can be fine-tuned in 2 days with 2 A100 GPU's using a small dataset of 323 videos. In contrast, PIDM \cite{pose-transfer-dm} is
        trained from scratch on 101,966 Deep Fashion \cite{deep-fashion} image pairs with 4 A100s for 26 days. 

    \subsection{Diffusion Models for Video Synthesis}
        Many text-to-video diffusion models rely on adapting text-to-image diffusion models for video synthesis \cite{flexible-diffusion-modeling,latent_vdm,imagen_video,video-diffusion-models,make-a-video,ddpm_video}. While the results are promising, these methods still struggle to match the realism that text-to-image models do. Quality is largely hindered due to the new challenges introduced by video synthesis, such as maintaining temporal consistency across frames and generating realistic motion. Some video diffusion methods are instead trained from scratch, requiring expensive computational resources, huge training datasets, and extensive training time~\cite{imagen_video,video-diffusion-models,dream_fusion,dalle2,dreamix, latent_vdm}. 
        Concurrently, Tune-A-Video fine-tunes a text-to-image pretrained diffusion model for text-and-image conditioned video generation \cite{tune-a-video}. However, like earlier video diffusion methods, Tune-A-Video's results exhibit textural flickering and structural inconsistencies. Our work aims to address these issues in order to synthesize realistic human and fabric motion.

    \subsection{Conditioning Mechanisms for Diffusion Models}
        Text-conditioning is popular among image diffusion models \cite{stable_diffusion,hierarchical_text_conditional,photorealistic_text2img_diffusion}. While effective at controlling high-level details, text conditioning fails to provide rich, detailed information about the \textit{exact} identity or pose of a person and garment. 
        
        Several works tackle the challenge of image conditioning for a pretrained text-to-image Stable Diffusion model~\cite{universal_guidance,instruct_pix2pix,encoder_personalization,dreamix,dreambooth,3dim}. 
        DreamBooth, the first method to perform subject-specific finetuning of Stable Diffusion on a set of images, learns a unique text token to represent the subject in the text encoder~\cite{dreambooth}. Others incorporate text to edit the appearance of existing images~\cite{instruct_pix2pix} and videos~\cite{tune-a-video,dreamix}. PIDM~\cite{pose-transfer-dm} encodes image textures using a separate textural encoder and concatenates target pose with an input noisy image.
        DreamPose allows the user to not only control the appearance of subjects in video, but also the structure and motion. Similar to PIDM, our image conditioning approach directly incorporates image embeddings in the cross-attention layers of the UNet, but these image embeddings come from a mixture of two pretrained encoders: CLIP and VAE. Moreover, with our method, we achieve smooth, temporally consistent motion using a multi-pose input representation concatenated to the input noise.
    
\section{Background}
    \textit{Diffusion models} are a recent class of generative models that have surpassed GANs at synthesis tasks in terms of quality, diversity, and training stability \cite{dms_beat_gans}. A standard image diffusion model learns to iteratively recover an image from normally distributed random noise \cite{diffusion_models}. A \textit{latent} diffusion model, e.g., Stable Diffusion~\cite{stable_diffusion}, operates in the encoded latent space of an autoencoder, thereby saving computational complexity, while sacrificing minimal perceptual quality.
    Stable Diffusion is composed of two models: a variational autoencoder and a denoising UNet. The autoencoder consists of an encoder $\mathcal{E}$ that distills a frame $x$ into a compact latent representation, $z = \mathcal{E}(x)$, and a decoder $\mathcal{D}$ that reconstructs the image from its latent representation, $x' = \mathcal{D}(z)$. 
    During training, the latent features $z$ are diffused in $T$ timesteps by a deterministic Gaussian process to produce noisy features $\tilde{z}_T$, indistinguishable from random noise. In order to recover the original image, a time-conditioned UNet is trained to iteratively predict the noise of the latent features corresponding to each timestep $t \in \{ 1, ..., T\}$. The UNet $\epsilon_{\theta}$ objective function is:

        \begin{equation}
            L_{DM} = \mathbb{E}_{z, \epsilon \in  \mathcal{N}(0,1)} [ || \epsilon - \epsilon_{\theta}(\tilde{z}_t, t, c)||^2_2 ]
        \end{equation}

    \noindent where $c$ represents the embeddings of conditional information, such as text, image, segmentation mask, etc. In the case of text-to-image Stable Diffusion, $c$ is obtained using a CLIP text encoder \cite{clip}.
    Finally, the predicted denoised latents $z'$ are decoded to recover the predicted image $x' = \mathcal{D}(x')$.

    \textit{Classifier-free guidance} is a mechanism in sampling that pushes the distribution of predicted noise towards the conditional distribution via an implicit classifier~\cite{classifier_free_guidance}. This is practically achieved by dropout, a training scheme that, with a random probability, replaces real conditioning inputs with null inputs ($\varnothing$). During inference, the conditional prediction is used to guide the unconditional prediction towards the conditional, using a guidance scalar weight $s$:

        \begin{equation}
            \epsilon_{\theta} = \epsilon_{\theta}(\tilde{z_t}, t, \varnothing) + s \cdot (\epsilon_{\theta}(\tilde{z_t}, t, c) - \epsilon_{\theta}(\tilde{z_t}, t, \varnothing))
        \end{equation}
        \label{eq:cfg}
    
\vspace{-2em}
\section{Method}
    Our method aims to produce photorealistic animated videos from a single image and a pose sequence. To achieve this, we fine-tune a pretrained Stable Diffusion model on a collection of fashion videos. This involves adapting the architecture of Stable Diffusion (which is a text-to-image model) to accept additional conditioning signals (image and pose), and to output temporally consistent content that can be viewed as a video.  

    In the coming section, we begin by  describing the architectural modifications in Section~\ref{sec:arch}. Then, we describe the two-stage finetuning strategy in Section~\ref{sec:ft}. Finally, in Section~\ref{sec:cfg}, we describe the inference process of generating an animated video from a still image, which involves a novel formulation of classifier-free guidance.

    \subsection{Overview}
        Given input image $x_0$ and poses $\{p_1, ..., p_N\}$, our method generates a video $\{x_1', ..., x_N'\}$, where $x_i'$ is the $i$-th predicted frame corresponding to input pose $p_i$. Our method relies on a pretrained latent diffusion model~\cite{stable_diffusion}, which is conditioned on an input image and a sequence of poses. At inference time, we generate each frame independently through a standard diffusion sampling procedure: starting with uniformly distributed Gaussian noise, the diffusion model is repeatedly queried with both conditioning signals to gradually denoise the noisy latent to a plausible estimate. Finally, the predicted denoised latent $z'_i$ is decoded to produce the predicted video frame $x'_i = \mathcal{D}(z'_i)$.
        
    \subsection{Architecture}
    \label{sec:arch}
        The DreamPose model is a pose- and image-conditioned image generation model that modifies and fine-tunes the original text-to-image Stable Diffusion model for the purpose of image animation. The objectives of image animation include: (1) faithfulness to the provided input image, (2) visual quality, and (3) temporal stability across generated frames. As such, DreamPose requires an image conditioning mechanism that captures the global structure, person identity, and fine-grained details of the garment, as well as a method to effectively condition the output image on target pose while also enabling temporal consistency between independently sampled output frames. We describe our approach to achieving these goals in the sections below. A diagram of our architecture can be found in Figure~\ref{fig:architecture}. Full implementation details are provided in the supplementary material.

        \begin{figure}[h!]
            \begin{center}
                \includegraphics[width=\linewidth]{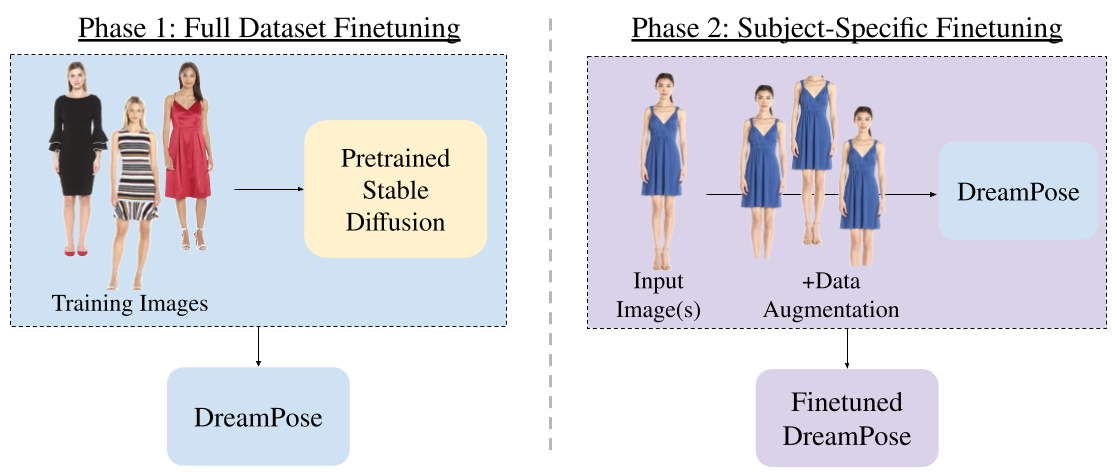}
            \end{center}
           \caption{Two-Phase Finetuning Scheme. In the first phase, our method fine-tunes the modified Stable Diffusion model on the full dataset. In the second phase, the model is further fine-tuned on a single subject image.}
        \label{fig:finetuning}
        \end{figure}
        
        \subsubsection{Split CLIP-VAE Encoder} In many prior works, such as InstructPix2Pix~\cite{instruct_pix2pix}, image conditioning signals are often concatenated with the input noise to the denoising U-Net. While this is effective for conditioning signals that are spatially aligned with the desired output image, in our case, our network aims specifically to produce images which are \emph{not} spatially aligned with the input image. As such, we explore alternative approaches for image conditioning. In particular, we implement image conditioning by replacing the CLIP text encoder with a custom conditioning adapter that combines the encoded information from pretrained CLIP image and VAE encoders.

        A crucial objective when finetuning from a pretrained network is to make training gradients as meaningful as possible by making the input signals as similar as possible to those used in the original network training. This helps avoid regressions in network performance during finetuning, or loss of learned priors, which can come from noisy gradients (e.g., if the network does not know how to parse new forms of input signals). For this reason, most diffusion-based finetuning schemes~\cite{instruct_pix2pix,zhang2023adding} will retain all original conditioning signals, and will initialize network weights that interact with new (previously unseen) conditioning signals to zero.  
        
        For our purposes, given that Stable Diffusion is conditioned on CLIP embeddings of text prompts, and CLIP encodes both text and images to a shared embedding space, it may seem natural to simply replace the CLIP conditioning with the embedding derived from the conditioning image. While this would in theory pose a very small change to the original architecture and allow for image conditioning with minimal finetuning, we find that in practice that CLIP image embeddings alone are insufficient for capturing fine-grained details in the conditioning image. So, we instead additionally input the encoded latent embeddings from Stable Diffusion's VAE. Adding these latent embeddings as conditioning has the added benefit of coinciding with the output domain of the diffusion model.

        Since the architecture does not support VAE latents as a conditioning signal by default, we add an adapter module $\mathcal{A}$ that combines the CLIP and VAE embeddings to produce one embedding that is used in the network's usual cross-attention operations. This adapter blends both the signals together and transforms the output into the typical shape expected by the cross-attention modules of the denoising U-Net. Initially, the weights corresponding to the VAE embeddings are set to zero, such that the network begins training with only the CLIP embeddings (as mentioned before, to mitigate network ``shock'' in training). We define the final image conditioning signal $c_I$ as:

            \begin{equation}
                c_I = \mathcal{A}(c_{\text{ CLIP}}, c_{\text{ VAE}})
            \end{equation}

        \def\Plus{\texttt{+}}
        \subsubsection{Modified UNet} Unlike the image conditioning, the pose conditioning \emph{is} image-aligned. As such, we concatenate the noisy latents $\tilde{z}_i$ with a target pose representation $c_p$. To account for noise in the poses (which are estimated from real videos using an off-the-shelf network~\cite{densepose}) and to maximize temporal consistency in the generated frames, we set $c_p$ to consist of five consecutive pose frames: $c_p = \{p_{i\mhyphen2}, p_{i\mhyphen1}, p_i, p_{i\Plus1}, p_{i\Plus2}\}$. We observe that individual poses are prone to frame-to-frame jitter, but training the network with a set of consecutive poses increases the overall motion smoothness and temporal consistency. Architecturally, we modify the UNet input layer to take in 10 extra input channels, initialized to zero, while the original channels corresponding to the noisy latents are unmodified from the pretrained weights. 
        
    \subsection{Finetuning}    
    \label{sec:ft}
        For initialization, the unmodified Stable Diffusion layers are initialized from a pretrained text-to-image Stable Diffusion checkpoint, except for the CLIP image encoder which is loaded from a separate pretrained checkpoint \cite{stable_diffusion,clip}. As mentioned previously, the novel layers are initialized such that initially the new conditioning signals do not contribute to the network output.
        
        Following initialization, DreamPose is fine-tuned in two stages (shown in Figure~\ref{fig:finetuning}). The first phase fine-tunes the UNet and adapter module on the full training dataset in order to synthesize frames consistent with an input image and pose. The second phase refines the base model by finetuning the UNet and adapter module, then the VAE decoder, on one or more subject-specific input image(s) to create a subject-specific custom model used for inference. 
        
        Similar to other image-conditional diffusion methods \cite{dreambooth,dream_fusion,dreamix}, we find that sample-specific finetuning is essential to preserving the identity of the input image's person and garment, as well as maintaining a consistent appearance across frames. However, simply training on a single frame and pose pair quickly leads to artifacts in the output videos, such as texture-sticking. To prevent this, we augment the image-and-pose pair at each step, such as by adding random cropping.
        
\begin{figure}[h!]
            \begin{center}
                \includegraphics[width=1.0\linewidth]{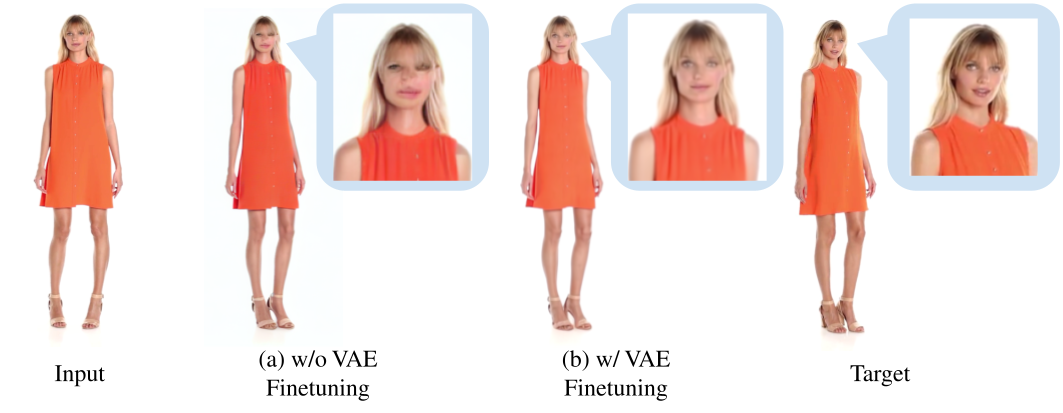}
            \end{center}
           \caption{Ablation of VAE Finetuning. We find that finetuning the VAE decoder, in addition to the UNet, during the subject-specific finetuning phase yields more photorealistic details and reduces high-frequency noise, compared to finetuning the UNet alone.}
        \label{fig:vae-finetuning}
        \end{figure}

        We also find that finetuning the VAE decoder is crucial for recovering sharper, more photorealistic details in the synthesized output frames. 
        Refer to Figure~\ref{fig:vae-finetuning} and the supplementary videos for an ablated comparison. Furthermore, we show in Figure~\ref{fig:dual-cfg} that even single image finetuning of the decoder allows increased pose guidance, without sacrificing the person identity or appearance.

        \begin{figure*}
            \begin{center}
                \includegraphics[width=0.8\linewidth]{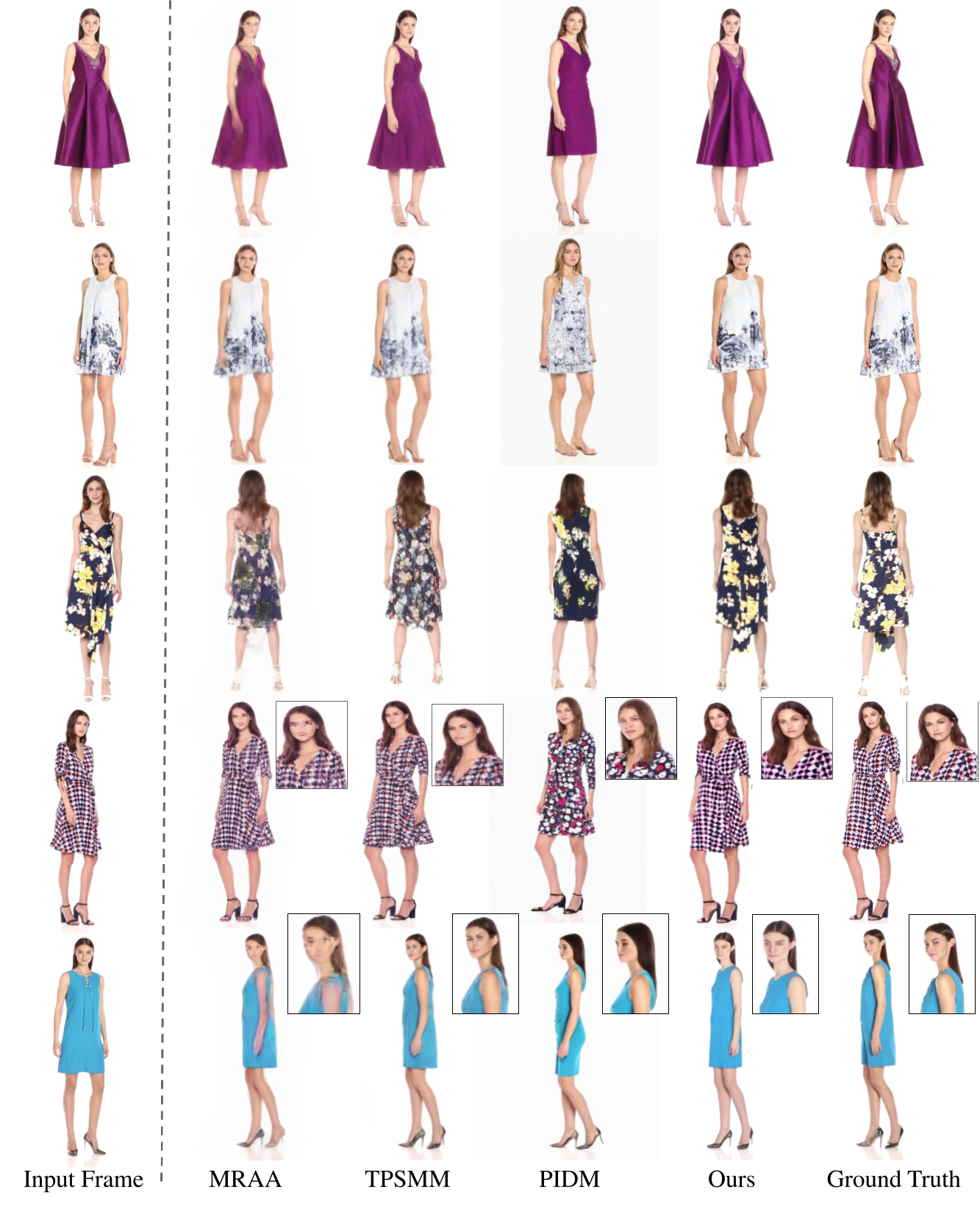}
            \end{center}
           \caption{Qualitative comparisons of our method versus MRAA~\cite{articulated_animation}, TPSMM~\cite{thin_plate_spline}, and PIDM~\cite{pose-transfer-dm}. Our method produces more photorealistic details aligned with the input frame, such as garment folds, fine-grain patterns, and face identity. Our method can also better handle challenging cases, including pattern synthesis in occluded regions.}
           \vspace{1em}
        \label{fig:qualitative}
        \end{figure*}
        
    \subsection{Pose and Image Classifier-Free Guidance}
    \label{sec:cfg}
        At inference time, we generate a video frame-by-frame from a single input image and a sequence of poses using the subject-specific model. We modulate the strength of image conditioning $c_I$ and pose conditioning $c_p$ during inference using dual classifier-free guidance \cite{instruct_pix2pix}. The dual classifier-free guidance equation is modified from Equation~\ref{eq:cfg} to be controlled by two guidance weights, $s_I$ and $s_p$, which rule how similar the output image is to the input image $c_I$ and input pose $c_p$, respectively:
        \begin{equation}
        \begin{aligned}
            \epsilon_{\theta}(z_t, c_I, c_p) & = \epsilon_{\theta}(z_t, \varnothing, \varnothing) \\
                            & + s_I (\epsilon_{\theta}(z_t, c_I, \varnothing) - \epsilon_{\theta}(z_t, \varnothing, \varnothing)) \\
                            & + s_p (\epsilon_{\theta}(z_t, c_I, c_p) - \epsilon_{\theta}(z_t, c_I, \varnothing))
        \end{aligned}
        \label{eq:cfg-equation}
        \end{equation}
        
\begin{figure}[t]
            \begin{center}
                \includegraphics[width=\linewidth]{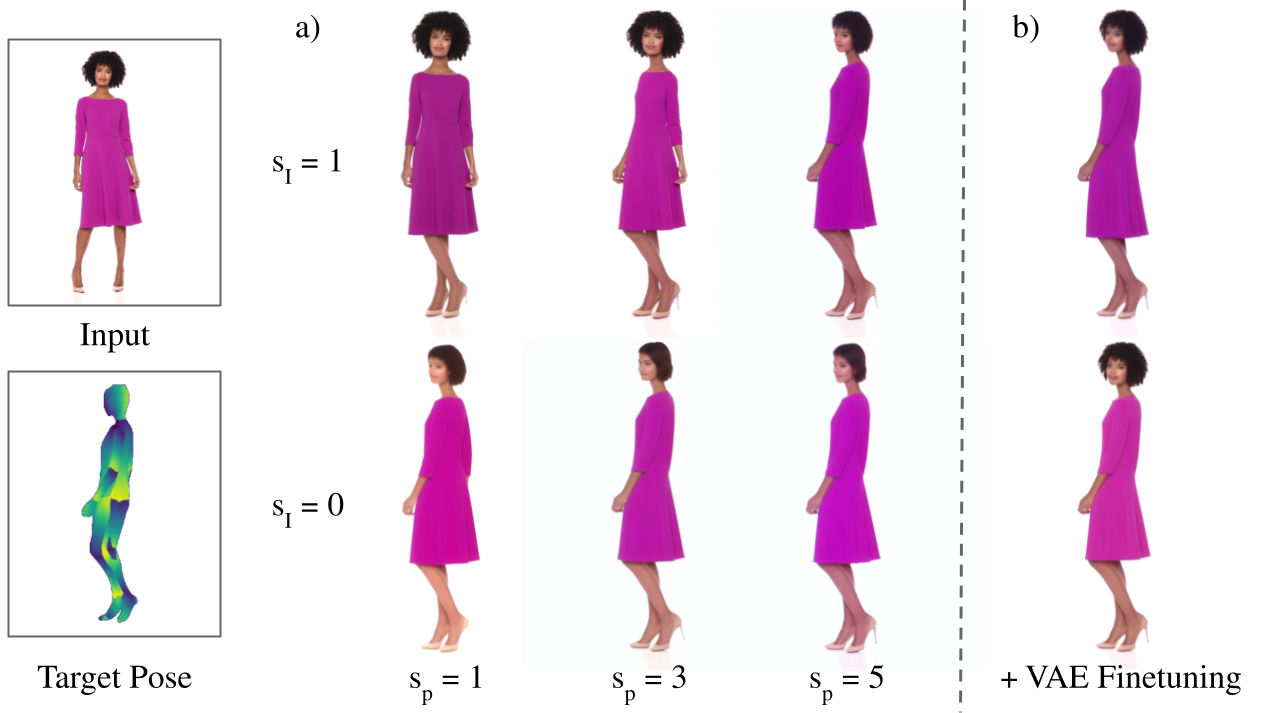}
            \end{center}
           \caption{ a) Pose and Image Classifier-Free Guidance. We demonstrate the effect of the relative weight between image and pose guidance weights, $s_I$ and $s_P$. Results shown are after subject-specific finetuning. b) VAE decoder finetuning improves the appearance and person identity, even with larger relative pose guidance. }
        \label{fig:dual-cfg}
        \end{figure}

        In Figure~\ref{fig:dual-cfg}, we show the effect of varying the classifier free guidance weights ($s_I, s_p$). A large $s_I$ ensures high appearance fidelity to the input image, while a large $s_p$ ensures alignment to the input pose. In addition to strengthening our pose and image guidance, the decoupled classifier-free guidance prevents overfitting to the one input pose after subject-specific finetuning.

    \subsection{Dataset}

    We train and test our method on the UBC Fashion dataset \cite{ubc_fashion}. We follow the provided train/test split of 339 training and 100 test videos. Each video has a frame rate of 30 frames/second and is approximately 12 seconds long. During training, we randomly sample pairs of frames from the training videos. We compute poses with DensePose \cite{densepose}.
        


        \begin{table*}[h!]
        \begin{center}
        \begin{tabular}{c c c c c c c c c}
        \hline
        & L1 $\downarrow$ & SSIM $\uparrow$ & VGG  $\downarrow$ & LPIPS  $\downarrow$ & FID $\downarrow$ & FVD (16f) $\downarrow$ & AED $\downarrow$ \\
        \hline\hline
        MRAA \cite{articulated_animation} &  0.0857 &  0.749 & 0.534 & 0.212 & 23.42 & 253.65 & 0.0139  \\
        TPSMM \cite{thin_plate_spline}  &  0.0858 & 0.746 & 0.547 & 0.213 & 22.87 & 247.55 & 0.0137 \\
        PIDM \cite{pose-transfer-dm} & 0.1098 & 0.713 & 0.629 & 0.288 & 30.279 & 1197.39 & 0.0155 \\
        Ours & \textbf{0.0256} & \textbf{0.885} & \textbf{0.235} & \textbf{0.068} & \textbf{13.04} & \textbf{238.75} & \textbf{0.0110}  \\
        \hline
        \end{tabular}
        \end{center}
        \caption{Quantitative comparisons of our method with MRAA, TPSMM, PIDM, and our method. Bolded values indicate best scores in each column. 
        }
        \label{quantitative-comparison}
        \end{table*}
\section{Results}
    DreamPose is capable of generating state-of-the-art fashion videos from still images. In Figure~\ref{fig:qualitative-results}, we showcase frames synthesized by DreamPose from a variety of input images and poses from the UBC Fashion dataset~\cite{ubc_fashion}. DreamPose handles diverse human and clothing appearances well, even from different viewpoints and loose garments. Moreover, for a given subject image, DreamPose can be conditioned on driving poses derived from a different video, as shown in the supplementary Figure~\ref{fig:diff-video-pose}. We also show additional results on images from the DeepFashion dataset~\cite{deep-fashion} in the supplementary materials.

    \subsection{Comparisons}
        We compare DreamPose quantitatively and qualitatively to two publicly available state-of-the-art conditional video synthesis methods, Motion Representations for Articulated Animation (MRAA)~\cite{articulated_animation} and Thin-Plate Spline Motion Model (TPSMM)~\cite{thin_plate_spline}, and a concurrent diffusion-based pose transfer method, PIDM~\cite{pose-transfer-dm}. We train all methods from scratch on the UBC Fashion Dataset~\cite{ubc_fashion}, using the provided training scripts and configurations. For evaluating MRAA~\cite{articulated_animation} and TPSMM~\cite{thin_plate_spline}, we use the provided test scripts in the ``AVD" mode. 
        
        \subsubsection{Quantitative Analysis} We present our quantitative analysis in Table~\ref{quantitative-comparison}. We test all models on the UBC Fashion test set, consisting of 100 unique fashion videos, at 256px resolution \cite{ubc_fashion}. For each video, we extract 50 frames for testing, where they are at least 50 frames away from the input frame. The full DreamPose model quantitatively outperforms all three methods in all metrics: L1, SSIM~\cite{ssim}, VGG~\cite{vgg}, LPIPS~\cite{lpips}, FID~\cite{fid}, FVD~\cite{FVD}, and Average Euclidean Distance (AED). Note that while PIDM produces realistic images, it struggles to preserve clothing and person identity, as indicated by the quantitative results. Moreover, there is poor temporal consistency frame-to-frame, resulting in a large FVD~\cite{FVD} score.

        \begin{figure*}[t]
            \begin{center}
                \includegraphics[width=1.0\linewidth]{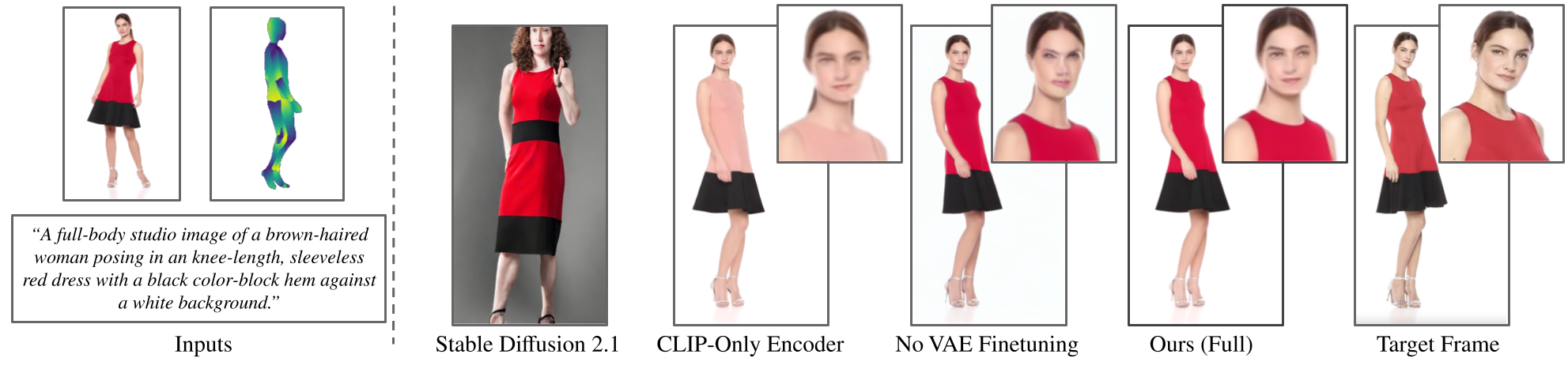}
            \end{center}
           \caption{Qualitative Ablation of Image Conditioning. We compare results of the original text-to-image Stable Diffusion model, our model with CLIP-only image embeddings without finetuning, our model with CLIP-VAE image encoder embeddings without VAE finetuning, and our full model.}
           \hspace{2em}
        \label{fig:ablations}
        \end{figure*}

        \subsubsection{Qualitative Analysis} We qualitatively compare our method to MRAA, TPSMM, and PIDM in Figure~\ref{fig:qualitative-results}.  With these other methods, note that the person identity, fabric folds, and fine patterns are lost in new poses, whereas DreamPose accurately retains those details. Plus, during large pose changes, MRAA may produce disjointed or blurry limbs. 
        Compared to PIDM, DreamPose produces higher-fidelity results, in terms of both face identity and clothing patterns. While PIDM synthesizes realistic faces, they do not necessarily align with the identity of the source person. Moreover, we find that both the identity and the dress appearance vary frame-to-frame. As such, PIDM does not work well as-is for video synthesis.  We provide additional comparisons to PIDM on the DeepFashion dataset \cite{deep-fashion} in the supplementary materials.
        Furthermore, we conduct a user study and provide the results in the supplementary material.
        
        

    \subsection{Ablation Studies}
    We perform comparisons of ablated versions of our method to verify our design choices. Namely, we compare five variants: \textbf{(1) Ours$_{\text{CLIP}}$:} We use a pretrained CLIP image encoder, instead of our dual CLIP-VAE encoder, \textbf{(2) Ours$_{\text{No-VAE-FT}}$:} We do subject-specific finetuning of the UNet only, not the VAE decoder, \textbf{(3) Ours$_{\text{1-pose}}$:} We concatenate only one target pose, instead of 5 consecutive poses, to the noise. \textbf{(4) Ours$_\text{smooth}$:} Like ablation Ours$_{\text{1-pose}}$, but apply temporal smoothing to the output frames. \textbf{(5) Ours$_\text{full}$:} Our full model, including subject-specific VAE finetuning, CLIP-VAE encoder, and 5-pose input.

    \paragraph{Quantitative Comparison.} For each ablated version, we compute the L1, SSIM, VGG, and LPIPS for 100 predicted video frames selected from each of the 100 test videos of the UBC Fashion dataset \cite{ubc_fashion}. Shown in Table~\ref{tab:quantitative-ablation}, our full model outperforms the ablated versions in all four metrics.

    \begin{table}[h!]
        \begin{center}
        \begin{tabular}{c c c c c c }
        \hline
        & L1 $\downarrow$ & SSIM $\uparrow$ & VGG  $\downarrow$ & LPIPS  $\downarrow$  \\
        \hline\hline
        Ours$_{\text{CLIP}}$ & 0.025 & 0.882 & 0.247 & 0.070 \\
        Ours$_{\text{No-VAE-FT}}$ & 0.025 & 0.897 & 0.210 & 0.057 \\
        Ours$_{\text{1-pose}}$ & 0.019 & 0.899 & 0.208 & 0.056\\
        Ours$_{\text{smooth}}$ & 0.767 & 0.758 & 0.502 & 0.202\\
        Ours$_{\text{full}}$ & \textbf{0.019} & \textbf{0.900} & \textbf{0.207} & \textbf{0.056} \\
        \hline
        \end{tabular}
        \end{center}
        \caption{Quantitative comparison of ablated versions of our method. Note that the single-pose version of our method, although achieving similar numerical results, fails to achieve the motion smoothness as the full method. Please refer to our project page for a video comparison.
        }
        \label{tab:quantitative-ablation}
        \end{table}

    \paragraph{Qualitative Comparison.} We visually show the effectiveness of our full method in Figure~\ref{fig:ablations}. We compare results from the original text-conditioned Stable Diffusion model, our method with only a CLIP image encoder, our method with CLIP-VAE encoder, and our full method with CLIP-VAE encoder and subject-specific VAE finetuning. 
    
    The original Stable Diffusion model with text-only conditioning via CLIP text encoder is unable to preserve rich details of the garment or person identity. Simply replacing the text encoder with a CLIP image encoder helps capture most image details, but there is still information loss about the appearance. Subject-specific finetuning of the UNet, similar to DreamBooth \cite{dreambooth}, is critical to preserving photorealistic details in the face and garment. Furthermore, we find that also finetuning the VAE decoder on the input image greatly improves the sharpness of these details and does not lead to overfitting to the input pose. 
    
    Lastly, with only a single input pose, there is noticeable flickering of the subject's shape, especially around the feet and hair. Please refer to video qualitative comparisons of each ablated version on our project page.
        \begin{figure}
            \begin{center}
                \includegraphics[width=\linewidth, height=8cm]{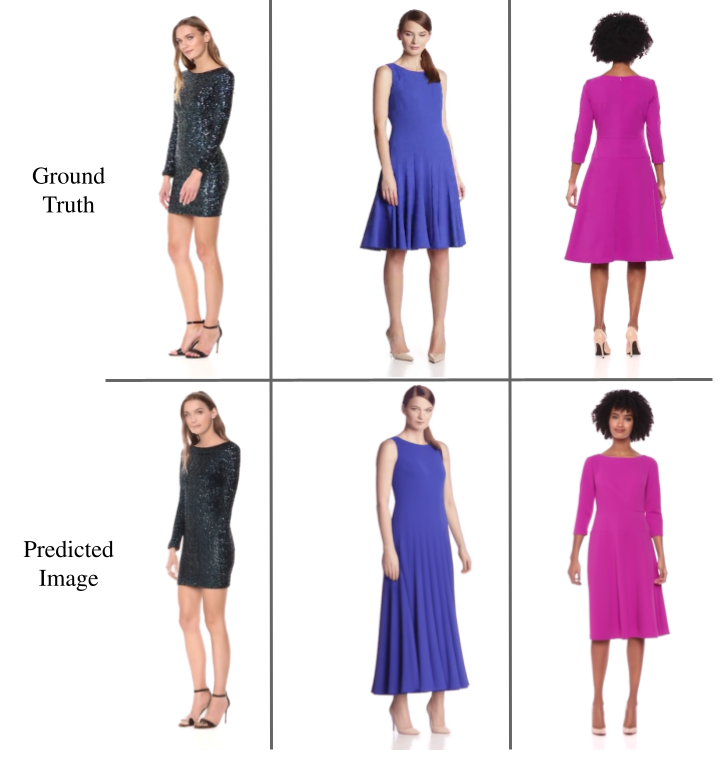}
            \end{center}
           \caption{Examples of failure cases of our method. Our method may merge limbs into underlying fabric textures (left), hallucinate features (middle), or predict front-facing person instead of a back-facing person (right).}
        \label{fig:failure-cases}
        \end{figure}    
\section{Limitations \& Future Work}  In Figure~\ref{fig:failure-cases}, we show failure cases of our method. On rare occasions, we observe limbs disappearing into the fabric, hallucinated dress features, and directional misalignment when the target pose is facing backwards. We suspect that some of these failures could be alleviated with improved pose estimation, a larger dataset, or a segmentation mask. Additionally, while our method produces realistic results on most plain and simple-patterned fabrics, some of our results present minor flickering behavior on large and complex patterns. Achieving better temporal consistency on such patterns, ideally without subject-specific finetuning, is left to future work. Lastly, similar to other diffusion models, our finetuning and inference times are slow compared to GAN or VAE methods. finetuning the model on a specific subject takes approximately 10 minutes for the UNet and 20 minutes for the VAE decoder, in addition to an 18 second per-frame rendering time. 

\section{Conclusion}
    In this paper, we presented DreamPose, a novel diffusion-based method for still fashion image animation. Given a single image and pose sequence, we demonstrate how our method generates photorealistic fashion videos from only a single image -- animating a diverse range of fabrics, patterns, and person identities. 

\section*{Acknowledgments}
\noindent This work was supported by NVIDIA and the UW Reality Lab, Meta, Google, OPPO, and Amazon.

{\small
\bibliographystyle{ieee_fullname}
\bibliography{arxiv}
}

\pagebreak

\onecolumn
\setcounter{section}{0}
\begin{center}
\textbf{\Large Supplementary Material}
\end{center}
\section{Implementation Details}
    Our experiments are trained on two NVIDIA A100 GPU's with resolution 512x512. In our first phase of training, we fine-tune our base model UNet on the full training dataset for a total of 5 epochs at a learning rate of $5e{\text{-}6}$. We use an effective batch size of 16 (through 4 gradient accumulation steps). We implement a dropout scheme where null values replace the pose input 5\% of the time, the input image 5\% of the time, and both input pose and input image 5\% of the time during training.
    We further fine-tune the UNet on a specific sample frame for another 500 steps with a learning rate of $1e{\text{-}5}$ and no dropout. Lastly, we fine-tune the VAE decoder only for 1500 steps with a learning rate of $5e{\text{-}5}$. 
    During inference, we use a PNDM sampler for 100 denoising steps \cite{pndm}.
    
\section{User Studies} 
    We conducted two user studies involving 50 distinct Amazon Mechanical Turk workers to compare our method with state-of-the art image animation approaches \cite{articulated_animation} \cite{thin_plate_spline} and evaluate the quality of our videos. In both surveys, workers evaluated results corresponding to 50 unique input images from the test set of the UBC Fashion dataset \cite{ubc_fashion}.
    
    In the first user study, workers were asked their pair-wise preferences between our method and one of the other methods. For each input image, the workers were shown two videos: one containing the input image, our resulting video, and the MRAA resulting video and the other containing the input image, our resulting video, and the TPSMM resulting video. The ordering of our video and other video (MRAA or TPSMM) was randomized for each question. For each videos, workers selected their preference between the videos. The results are shown in Table~\ref{user-survey}. Overall, the workers had a preference for our method over MRAA and TPSMM.
    
\begin{figure*}[h!]
        \begin{center}
            \includegraphics[width=0.5\linewidth]{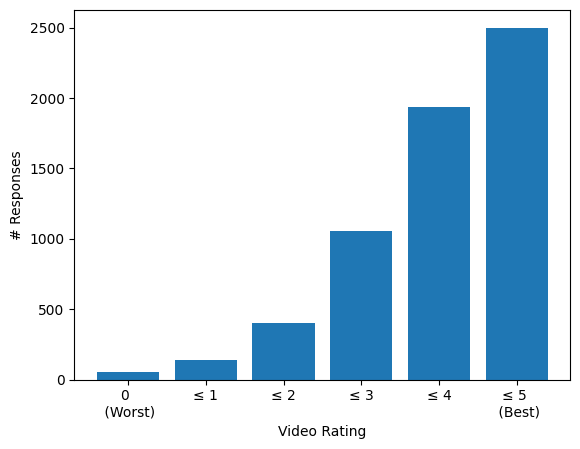}
        \end{center}
       \caption{ Results of User Study \#2: Amazon Mechanical Turk worker ratings of our videos from 0 (video does not match input image) to 5 (video is a realistic animation of the input image). Overall, $85\%$ of workers rated our method a 3 or higher.}
    \label{fig:user-ratings}
    \end{figure*}

    In the second user study, workers were asked to rate our videos and TPSMM videos on a scale of 0 to 5, where 0 corresponds a video that does not match the input image at all and 5 corresponds to a realistic animation of the input image. During training, workers were shown a video of a different dress for as an example of a "0" rating and a ground-truth video of the input image as an example of the "5" rating. The results are shown in Figure~\ref{fig:user-ratings}. Our videos achieved higher scores for image similarity and quality than TPSMM and ${85\%}$ of users rated the results of our method a 3 or higher.
    
    \begin{table}[h!]
        \begin{center}
        \begin{tabular}{c c c c }
        \hline
        & $\#$ Responses & Total Responses & ($\%$) \\
        \hline\hline
        Ours $>$ MRAA \cite{articulated_animation} & 1637 & 2500 & (65\%)\\
        Ours $>$ TPSMM \cite{thin_plate_spline} & 1417 & 2500 & (57\%) \\
        \hline
        \end{tabular}
        \end{center}
        \caption{Results of User Study \#1: Workers choose between pairs of videos corresponding to input images, either our result vs. MRAA result or our result vs. TPSMM result. Overall, participants preferred our method over both MRAA and TPSMM in terms of quality and similarity to the input image.}
        \label{user-survey}
    \end{table}

\section{Different Videos for Source Person and Driving Pose Sequence}
\begin{figure}[h!]
            \begin{center}
                \includegraphics[width=0.8\linewidth]{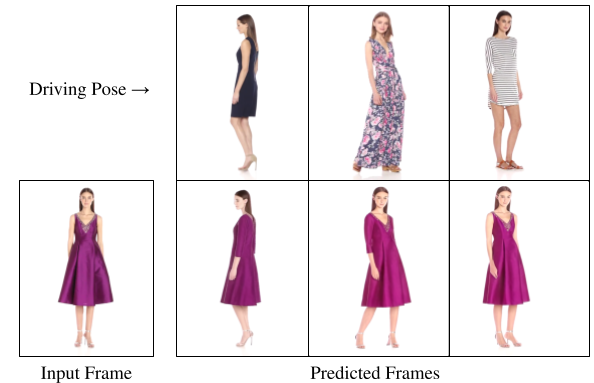}
            \end{center}
        \caption{Qualitative results for conditioning on subject and pose from different videos.}
        \label{fig:diff-video-pose}
        \end{figure}

    We show in Figure~\ref{fig:diff-video-pose} that DreamPose can animate an input image using motion from a video containing a different person and garment identity. As such, our method is applicable in practice when ground-truth motion is unavailable.

\section{Multiple Input Frames}
\begin{figure}
            \begin{center}
                \includegraphics[width=0.8\linewidth]{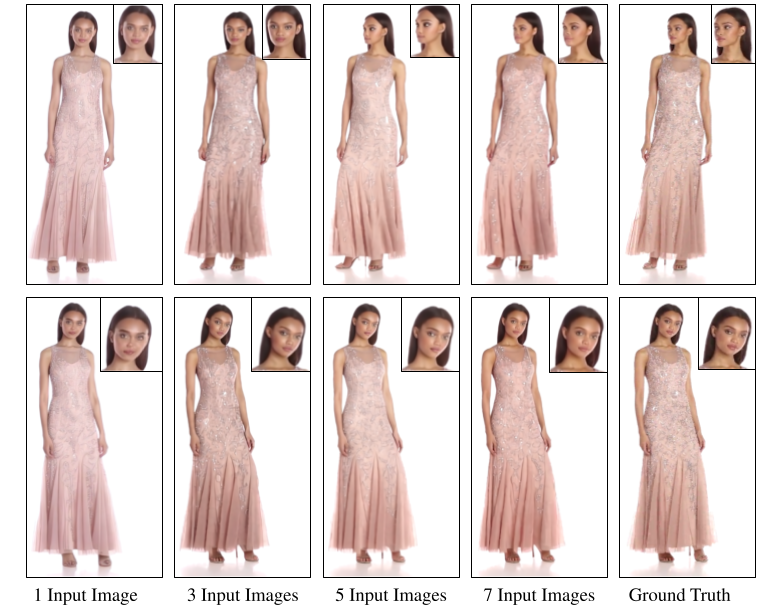}
            \end{center}
           \caption{Results after training with 1, 3, 5, and 7 input images. Increasing the number of input frames improves fidelity of pose, facial identity, and color.  }
        \label{fig:multi-image}
        \end{figure}

    While DreamPose demonstrates high-quality results with only a single input image, DreamPose can also be fine-tuned with an arbitrary number of input images of a subject. We showcase the results of training with multiple input images in Figure~\ref{fig:multi-image}. We find that additional input images of a subject increase the quality and viewpoint consistency.

\section{Deep Fashion Results} 
\begin{figure*}[h!]
            \begin{center}
                \includegraphics[width=1.0\linewidth]{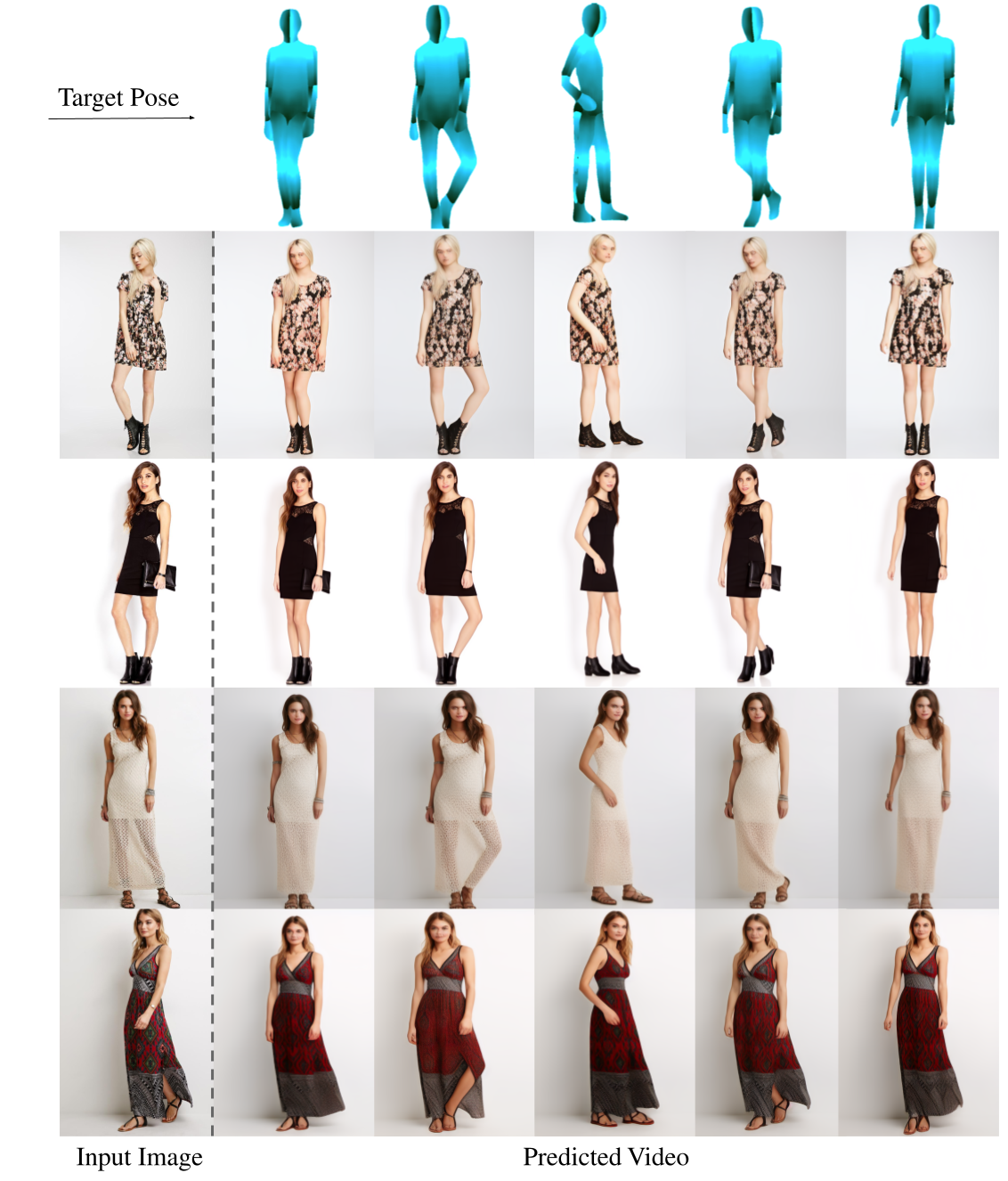}
            \end{center}
           \caption{ DreamPose results on unseen samples from the DeepFashion dataset \cite{deep-fashion}. Despite being trained exclusively on the UBC Fashion Dataset, our method generalizes to new garments and model identities after subject-specific finetuning of the base model.}
        \label{fig:deep-fashion}
        \end{figure*}

    We demonstrate the effectiveness of our method on a popular dataset, DeepFashion, in Figure~\ref{fig:deep-fashion} \cite{pose-transfer-dm, deep-fashion}. Although trained exclusively on the UBC Fashion video dataset, DreamPose performs well on unseen retail images, even to new backgrounds, model identities, accessories, and patterns.

\section{Application to Pose Transfer}
\begin{figure*}[h!]
            \begin{center}
                \includegraphics[width=1.0\linewidth]{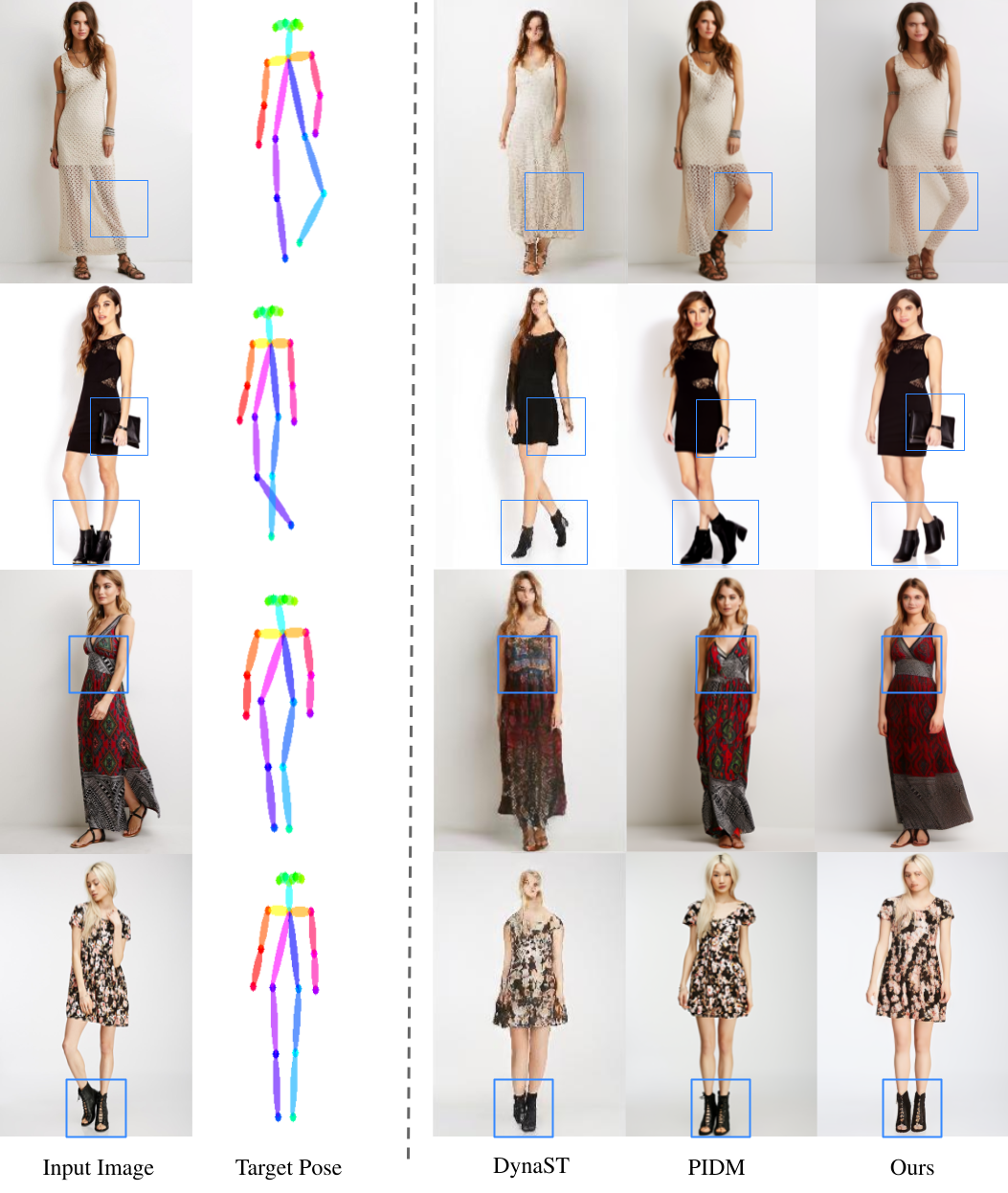}
            \end{center}
           \caption{Comparison of Pose Transfer Results. We compare our method to two state-of-the-art pose transfer methods, DynaST \cite{DynaST} and PIDM \cite{pose-transfer-dm}.}
        \label{fig:pose-transfer}
        \end{figure*}

    While adapted for image-to-video synthesis, DreamPose is also an effective pose transfer tool. In Figure~\ref{fig:pose-transfer}, we compare DreamPose to two state-of-the-art pose transfer models: DynaST \cite{DynaST} and PIDM \cite{pose-transfer-dm}. Our method is better able to preserve fine-details, such as shoe appearance, hemline, and face identity, than DynaST or PIDM. 

\end{document}

%% file: Figures/teaser.tex
\vspace{-3em}
\begin{center}%
    \captionsetup{type=figure}%
    \includegraphics[width=1.0\linewidth]{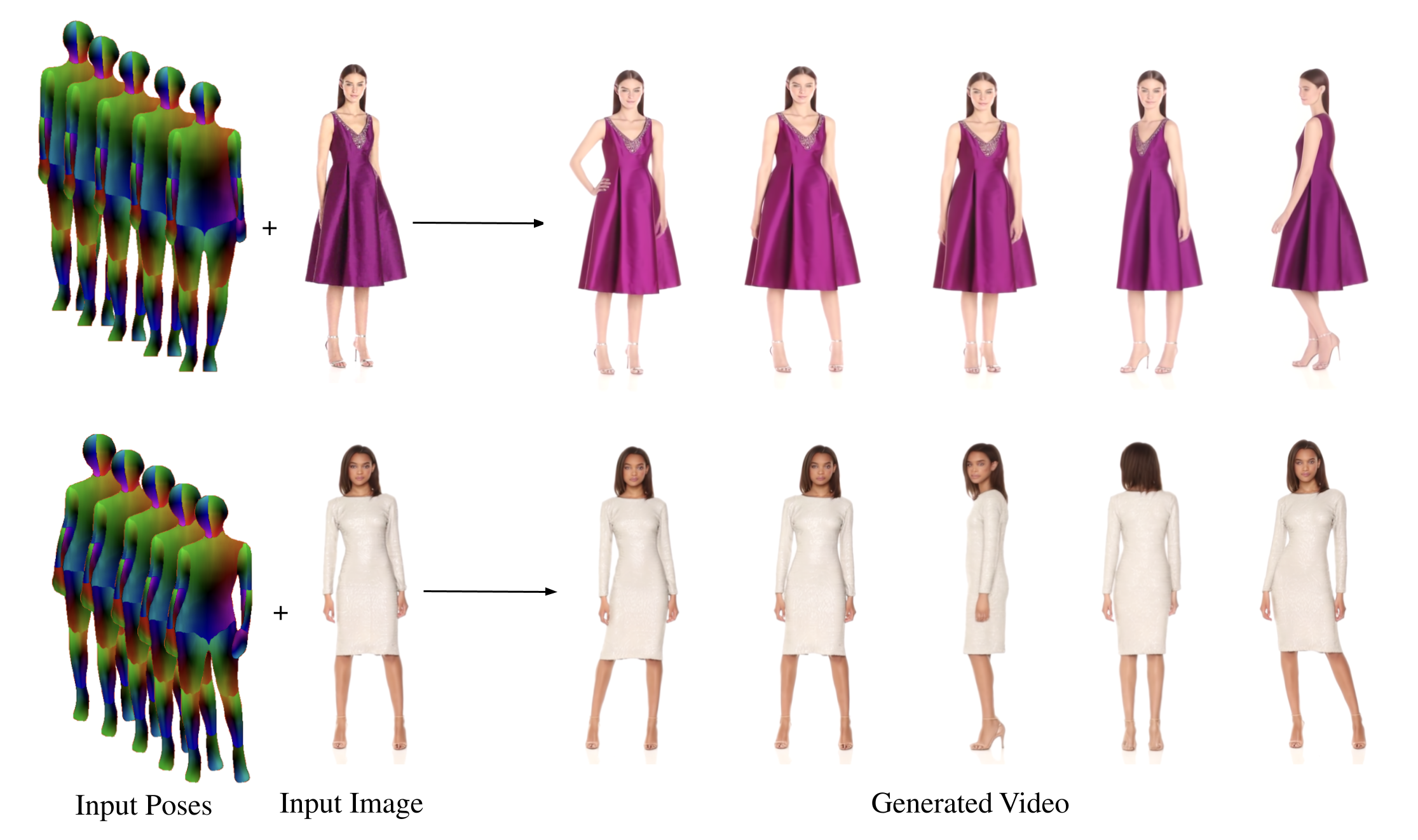}%
    \vspace{1em}
    \captionof{figure}{Given an image of a person and a sequence of body poses, DreamPose synthesizes a photorealistic video.
     }%
    \label{fig:teaser}%
\end{center}%
\vspace{1em}

%% file: arxiv.bbl
\begin{thebibliography}{10}\itemsep=-1pt

\bibitem{pose_with_style}
Badour AlBahar, Jingwan Lu, Jimei Yang, Zhixin Shu, Eli Shechtman, and Jia-Bin
  Huang.
\newblock Pose with style: Detail-preserving pose-guided image synthesis with
  conditional stylegan, 2021.

\bibitem{universal_guidance}
Arpit Bansal, Hong-Min Chu, Avi Schwarzschild, Soumyadip Sengupta, Micah
  Goldblum, Jonas Geiping, and Tom Goldstein.
\newblock Universal guidance for diffusion models, 2023.

\bibitem{dain}
Wenbo Bao, Wei-Sheng Lai, Chao Ma, Xiaoyun Zhang, Zhiyong Gao, and Ming-Hsuan
  Yang.
\newblock Depth-aware video frame interpolation, 2019.

\bibitem{pose-transfer-dm}
Ankan~Kumar Bhunia, Salman Khan, Hisham Cholakkal, Rao~Muhammad Anwer, Jorma
  Laaksonen, Mubarak Shah, and Fahad~Shahbaz Khan.
\newblock Person image synthesis via denoising diffusion model, 2022.

\bibitem{instruct_pix2pix}
Tim Brooks, Aleksander Holynski, and Alexei~A. Efros.
\newblock Instructpix2pix: Learning to follow image editing instructions, 2022.

\bibitem{DiffFashion}
Shidong Cao, Wenhao Chai, Shengyu Hao, Yanting Zhang, Hangyue Chen, and Gaoang
  Wang.
\newblock Difffashion: Reference-based fashion design with structure-aware
  transfer by diffusion models, 2023.

\bibitem{viton_hd}
Seunghwan Choi, Sunghyun Park, Minsoo Lee, and Jaegul Choo.
\newblock Viton-hd: High-resolution virtual try-on via misalignment-aware
  normalization, 2021.

\bibitem{DiOr}
Aiyu Cui, Daniel McKee, and Svetlana Lazebnik.
\newblock Dressing in order: Recurrent person image generation for pose
  transfer, virtual try-on and outfit editing, 2021.

\bibitem{dms_beat_gans}
Prafulla Dhariwal and Alex Nichol.
\newblock Diffusion models beat gans on image synthesis, 2021.

\bibitem{encoder_personalization}
Rinon Gal, Moab Arar, Yuval Atzmon, Amit~H. Bermano, Gal Chechik, and Daniel
  Cohen-Or.
\newblock Designing an encoder for fast personalization of text-to-image
  models, 2023.

\bibitem{deep-fashion}
Yuying Ge, Ruimao Zhang, Lingyun Wu, Xiaogang Wang, Xiaoou Tang, and Ping Luo.
\newblock Deepfashion2: A versatile benchmark for detection, pose estimation,
  segmentation and re-identification of clothing images, 2019.

\bibitem{densepose}
Rıza~Alp Güler, Natalia Neverova, and Iasonas Kokkinos.
\newblock Densepose: Dense human pose estimation in the wild, 2018.

\bibitem{flexible-diffusion-modeling}
William Harvey, Saeid Naderiparizi, Vaden Masrani, Christian Weilbach, and
  Frank Wood.
\newblock Flexible diffusion modeling of long videos, 2022.

\bibitem{latent_vdm}
Yingqing He, Tianyu Yang, Yong Zhang, Ying Shan, and Qifeng Chen.
\newblock Latent video diffusion models for high-fidelity video generation with
  arbitrary lengths, 2022.

\bibitem{fid}
Martin Heusel, Hubert Ramsauer, Thomas Unterthiner, Bernhard Nessler, and Sepp
  Hochreiter.
\newblock Gans trained by a two time-scale update rule converge to a local nash
  equilibrium.
\newblock 2017.

\bibitem{stable_diffusion}
Jonathan Ho, William Chan, Chitwan Saharia, Jay Whang, Ruiqi Gao, Alexey
  Gritsenko, Diederik~P. Kingma, Ben Poole, Mohammad Norouzi, David~J. Fleet,
  and Tim Salimans.
\newblock Imagen video: High definition video generation with diffusion models,
  2022.

\bibitem{imagen_video}
Jonathan Ho, William Chan, Chitwan Saharia, Jay Whang, Ruiqi Gao, Alexey
  Gritsenko, Diederik~P. Kingma, Ben Poole, Mohammad Norouzi, David~J. Fleet,
  and Tim Salimans.
\newblock Imagen video: High definition video generation with diffusion models,
  2022.

\bibitem{classifier_free_guidance}
Jonathan Ho and Tim Salimans.
\newblock Classifier-free diffusion guidance, 2022.

\bibitem{video-diffusion-models}
Jonathan Ho, Tim Salimans, Alexey Gritsenko, William Chan, Mohammad Norouzi,
  and David~J. Fleet.
\newblock Video diffusion models, 2022.

\bibitem{eulerian_motion}
Aleksander Holynski, Brian Curless, Steven~M. Seitz, and Richard Szeliski.
\newblock Animating pictures with eulerian motion fields, 2020.

\bibitem{eva3d}
Fangzhou Hong, Zhaoxi Chen, Yushi Lan, Liang Pan, and Ziwei Liu.
\newblock Eva3d: Compositional 3d human generation from 2d image collections,
  2022.

\bibitem{vgg}
Justin Johnson, Alexandre Alahi, and Li Fei-Fei.
\newblock Perceptual losses for real-time style transfer and super-resolution,
  2016.

\bibitem{try_on_gan}
Kathleen~M Lewis, Srivatsan Varadharajan, and Ira Kemelmacher-Shlizerman.
\newblock Tryongan: Body-aware try-on via layered interpolation, 2021.

\bibitem{pndm}
Luping Liu, Yi Ren, Zhijie Lin, and Zhou Zhao.
\newblock Pseudo numerical methods for diffusion models on manifolds, 2022.

\bibitem{DynaST}
Songhua Liu, Jingwen Ye, Sucheng Ren, and Xinchao Wang.
\newblock Dynast: Dynamic sparse transformer for exemplar-guided image
  generation, 2022.

\bibitem{implicit_warping}
Arun Mallya, Ting-Chun Wang, and Ming-Yu Liu.
\newblock Implicit warping for animation with image sets, 2022.

\bibitem{dreamix}
Eyal Molad, Eliahu Horwitz, Dani Valevski, Alex~Rav Acha, Yossi Matias, Yael
  Pritch, Yaniv Leviathan, and Yedid Hoshen.
\newblock Dreamix: Video diffusion models are general video editors, 2023.

\bibitem{dream_fusion}
Ben Poole, Ajay Jain, Jonathan~T. Barron, and Ben Mildenhall.
\newblock Dreamfusion: Text-to-3d using 2d diffusion, 2022.

\bibitem{clip}
Alec Radford, Jong~Wook Kim, Chris Hallacy, Aditya Ramesh, Gabriel Goh,
  Sandhini Agarwal, Girish Sastry, Amanda Askell, Pamela Mishkin, Jack Clark,
  Gretchen Krueger, and Ilya Sutskever.
\newblock Learning transferable visual models from natural language
  supervision, 2021.

\bibitem{hierarchical_text_conditional}
Aditya Ramesh, Prafulla Dhariwal, Alex Nichol, Casey Chu, and Mark Chen.
\newblock Hierarchical text-conditional image generation with clip latents,
  2022.

\bibitem{dalle2}
Royi Rassin, Shauli Ravfogel, and Yoav Goldberg.
\newblock Dalle-2 is seeing double: Flaws in word-to-concept mapping in
  text2image models, 2022.

\bibitem{film}
Fitsum Reda, Janne Kontkanen, Eric Tabellion, Deqing Sun, Caroline Pantofaru,
  and Brian Curless.
\newblock Film: Frame interpolation for large motion, 2022.

\bibitem{global_flow_local_attn}
Yurui Ren, Xiaoming Yu, Junming Chen, Thomas~H. Li, and Ge Li.
\newblock Deep image spatial transformation for person image generation, 2020.

\bibitem{dreambooth}
Nataniel Ruiz, Yuanzhen Li, Varun Jampani, Yael Pritch, Michael Rubinstein, and
  Kfir Aberman.
\newblock Dreambooth: Fine tuning text-to-image diffusion models for
  subject-driven generation, 2022.

\bibitem{imagen}
Chitwan Saharia, William Chan, Saurabh Saxena, Lala Li, Jay Whang, Emily
  Denton, Seyed Kamyar~Seyed Ghasemipour, Burcu~Karagol Ayan, S.~Sara Mahdavi,
  Rapha~Gontijo Lopes, Tim Salimans, Jonathan Ho, David~J Fleet, and Mohammad
  Norouzi.
\newblock Photorealistic text-to-image diffusion models with deep language
  understanding, 2022.

\bibitem{photorealistic_text2img_diffusion}
Chitwan Saharia, William Chan, Saurabh Saxena, Lala Li, Jay Whang, Emily
  Denton, Seyed Kamyar~Seyed Ghasemipour, Burcu~Karagol Ayan, S.~Sara Mahdavi,
  Rapha~Gontijo Lopes, Tim Salimans, Jonathan Ho, David~J Fleet, and Mohammad
  Norouzi.
\newblock Photorealistic text-to-image diffusion models with deep language
  understanding, 2022.

\bibitem{monkey-net}
Aliaksandr Siarohin, Stéphane Lathuilière, Sergey Tulyakov, Elisa Ricci, and
  Nicu Sebe.
\newblock Animating arbitrary objects via deep motion transfer, 2018.

\bibitem{fomm}
Aliaksandr Siarohin, Stéphane Lathuilière, Sergey Tulyakov, Elisa Ricci, and
  Nicu Sebe.
\newblock First order motion model for image animation, 2020.

\bibitem{articulated_animation}
Aliaksandr Siarohin, Oliver~J. Woodford, Jian Ren, Menglei Chai, and Sergey
  Tulyakov.
\newblock Motion representations for articulated animation.
\newblock 2021.

\bibitem{make-a-video}
Uriel Singer, Adam Polyak, Thomas Hayes, Xi Yin, Jie An, Songyang Zhang, Qiyuan
  Hu, Harry Yang, Oron Ashual, Oran Gafni, Devi Parikh, Sonal Gupta, and Yaniv
  Taigman.
\newblock Make-a-video: Text-to-video generation without text-video data, 2022.

\bibitem{make-a-video-3d}
Uriel Singer, Shelly Sheynin, Adam Polyak, Oron Ashual, Iurii Makarov, Filippos
  Kokkinos, Naman Goyal, Andrea Vedaldi, Devi Parikh, Justin Johnson, and Yaniv
  Taigman.
\newblock Text-to-4d dynamic scene generation, 2023.

\bibitem{diffusion_models}
Jascha Sohl-Dickstein, Eric~A. Weiss, Niru Maheswaranathan, and Surya Ganguli.
\newblock Deep unsupervised learning using nonequilibrium thermodynamics, 2015.

\bibitem{FVD}
Thomas Unterthiner, Sjoerd van Steenkiste, Karol Kurach, Raphael Marinier,
  Marcin Michalski, and Sylvain Gelly.
\newblock Towards accurate generative models of video: A new metric \&
  challenges, 2018.

\bibitem{latent_image_animator}
Yaohui Wang, Di Yang, Francois Bremond, and Antitza Dantcheva.
\newblock Latent image animator: Learning to animate images via latent space
  navigation, 2022.

\bibitem{ssim}
Zhou Wang, A.C. Bovik, H.R. Sheikh, and E.P. Simoncelli.
\newblock Image quality assessment: from error visibility to structural
  similarity.
\newblock {\em IEEE Transactions on Image Processing}, 13(4):600--612, 2004.

\bibitem{3dim}
Daniel Watson, William Chan, Ricardo Martin-Brualla, Jonathan Ho, Andrea
  Tagliasacchi, and Mohammad Norouzi.
\newblock Novel view synthesis with diffusion models, 2022.

\bibitem{photo_wake-up}
Chung-Yi Weng, Brian Curless, and Ira Kemelmacher-Shlizerman.
\newblock Photo wake-up: 3d character animation from a single photo, 2018.

\bibitem{tune-a-video}
Jay~Zhangjie Wu, Yixiao Ge, Xintao Wang, Weixian Lei, Yuchao Gu, Wynne Hsu,
  Ying Shan, Xiaohu Qie, and Mike~Zheng Shou.
\newblock Tune-a-video: One-shot tuning of image diffusion models for
  text-to-video generation, 2022.

\bibitem{ddpm_video}
Ruihan Yang, Prakhar Srivastava, and Stephan Mandt.
\newblock Diffusion probabilistic modeling for video generation, 2022.

\bibitem{ubc_fashion}
Polina Zablotskaia, Aliaksandr Siarohin, Bo Zhao, and Leonid Sigal.
\newblock Dwnet: Dense warp-based network for pose-guided human video
  generation, 2019.

\bibitem{pise}
Jinsong Zhang, Kun Li, Yu-Kun Lai, and Jingyu Yang.
\newblock Pise: Person image synthesis and editing with decoupled gan, 2021.

\bibitem{zhang2023adding}
Lvmin Zhang and Maneesh Agrawala.
\newblock Adding conditional control to text-to-image diffusion models.
\newblock {\em arXiv preprint arXiv:2302.05543}, 2023.

\bibitem{lpips}
Richard Zhang, Phillip Isola, Alexei~A. Efros, Eli Shechtman, and Oliver Wang.
\newblock The unreasonable effectiveness of deep features as a perceptual
  metric, 2018.

\bibitem{thin_plate_spline}
Jian Zhao and Hui Zhang.
\newblock Thin-plate spline motion model for image animation, 2022.

\bibitem{progressive_pose_transfer}
Zhen Zhu, Tengteng Huang, Baoguang Shi, Miao Yu, Bofei Wang, and Xiang Bai.
\newblock Progressive pose attention transfer for person image generation,
  2019.

\end{thebibliography}
